# Image Enhancement using Fuzzy Intensity Measure and Adaptive Clipping Histogram Equalization

Xiangyuan Zhu, Xiaoming Xiao, Tardi Tjahjadi, Zhihu Wu, Jin Tang*

*Abstract*—Image enhancement aims at processing an input image so that the visual content of the output image is more pleasing or more useful for certain applications. Although histogram equalization is widely used in image enhancement due to its simplicity and effectiveness, it changes the mean brightness of the enhanced image and introduces a high level of noise and distortion. To address these problems, this paper proposes image enhancement using fuzzy intensity measure and adaptive clipping histogram equalization (FIMHE). FIMHE uses fuzzy intensity measure to first segment the histogram of the original image, and then clip the histogram adaptively in order to prevent excessive image enhancement. Experiments on the Berkeley database and CVF-UGR-Image database show that FIMHE outperforms state-of-the-art histogram equalization based methods.

## I. INTRODUCTION

THE aim of image enhancement is to bring out the hidden details in a low contrast image [1]. Image enhancement has been widely used in medical imaging [2], face recognition [3], and underwater images [4]. One image enhancement method, histogram equalization (HE), is widely used for contrast enhancement due to its simple operation and effectiveness. The main idea of HE is to flatten the probability density function of the input image and remap the grey levels to generate a processed image with improved contrast [5]. Although HE has many advantages, it significantly modifies the average brightness of the processed image with respect to the original image. It also introduces noise and intensity saturation effects which result in a loss of image details and make the appearance of the processed image unnatural.

In this paper, a new method is proposed to address the above-mentioned problems of HE, and verify the effectiveness of the method. These are achieved by using a fuzzy intensity measure for segmenting the original image histogram and adaptive clipping of the histogram to prevent excessive enhancement.

This paper is organized as follows. Section II reviews related methods. Section III presents the proposed method. Section IV analyzes the experimental results of the performance of the proposed method, and finally Section V concludes the paper.

## II. RELATED WORK

Image enhancement algorithms based on the classical histogram equalization have been proposed to overcome the disadvantages of HE. The earliest improved algorithm is the brightness preserving bi-histogram equalization (BBHE) [6]. BBHE divides the histogram of an image into two parts based on the average brightness of the image, and then HE is applied to each part independently. Dualistic sub-image histogram equalization (DSIHE) [7] which is based on BBHE uses the median of the input image grey levels to divide the histogram. DSIHE is better than BBHE in maintaining the image brightness. The minimum mean brightness error bi-histogram equalization (MMBEBHE) [8] computes a minimum absolute value of the average luminance difference of the input image and uses each grey level as a splitting threshold until the minimum absolute mean brightness error is obtained.

Recursive mean-separate histogram equalization (RMSHE) [9] utilizes BBHE recursively, and the histogram of the input image is divided into $2^r$ parts, where the user-defined scale $r$ determines the recursive level, thus allowing brightness preservation from 0% to 100% [5]. Another algorithm similar to RMSHE is recursive sub-image histogram equalization (RSIHE) [10] which uses DSIHE instead of BBHE to divide the histogram. The common difficulty with RSMHE and RSIHE is in choosing the appropriate value of $r$.

Bi-histogram equalization with a plateau limit (BHEPL) [11] first divides an image histogram into two parts using the mean brightness of the image, and then clips the sub-histogram bins independently. A variant of BHEPL, bi-histogram equalization median plateau limit (BHEPL-D) [12], divides a histogram into two parts using the median of the image grey levels. BHEPL-D is better than BHEPL in maintaining the mean brightness while preserving the details of the image.

In modified histogram equalization (MHE) [13] the

Manuscript received September 26, 2018; revised January 10, 2019. This work was supported by the National Natural Science Foundation of China (Grant NO. 61502537) and Natural Science Foundation of Hunan Province, China (Grant NO. 2016JJ2150) and the major research plan integration project (Grant NO. 91220301).

Xiangyuan Zhu is with School of automation, Central South University, China (e-mail: zhuxiangyuan@csu.edu.cn).
Xiaoming Xiao is with School of automation, Central South University, China (e-mail: xmxiao@csu.edu.cn).
Tardi Tjahjadi is with School of Engineering, University of Warwick, United Kingdom (e-mail: T.Tjahjadi@warwick.ac.uk).
Zhihu Wu is with School of automation, Central South University, China (e-mail: wuzhihu@csu.edu.cn).
Jin Tang is with School of automation, Central South University, China (corresponding author, e-mail: tjin@csu.edu.cn).

average value of the histogram is set as a limit to the accumulation of histogram component bins. The method, exposure based sub image histogram equalization (ESIHE) [14], first clips the histogram bins and then divides the image into two parts using an exposure threshold. This method is effective in enhancing under-exposed images. More recently, bi-histogram equalization using two different plateau limits (BHE2PL) [15] uses two plateau limits to modify the sub-histograms. The method preserves the mean brightness while enhancing the contrast of the processed image.

Although many new methods have been proposed to solve the problem of image contrast enhancement while maintaining brightness, these methods can only achieve good enhancement under certain conditions and cannot achieve the best results on all evaluation metrics. In this paper, a method named fuzzy intensity measure and adaptive clipping histogram equalization (FIMHE) is proposed. The method increases the image contrast while preserving the original information of the image.

## III. PROPOSED METHOD

FIMHE consists of four steps: computation of fuzzy intensity measure, histogram segmentation, adaptive histogram clipping and equalization.

### A. Fuzzy Intensity Measure

The fuzzy intensity measure [16] is defined to distinguish the bright and dark regions in an image. The mean and standard deviation of the histogram intensity distribution, i.e., respectively

$$g_a = \frac{\sum_{m=0}^{L-1} m \times p(m)}{\sum_{m=0}^{L-1} p(m)} \quad (1)$$

$$g_d = \left[\frac{\sum_{m=0}^{L-1}\left[(m-g_a)^2 p(m)\right]}{\sum_{m=0}^{L-1} p(m)}\right]^{1/2} \quad (2)$$

are used to determine the non-homogeneous intensity distribution of the image. They are also used to define the

$$\text{fuzzy intensity measure} = \frac{g_d}{g_a} \quad (3)$$

and

$$T = L\left[\frac{g_d}{g_a}\right], \quad (4)$$

where $m$ represents the intensity of the pixel at image coordinates $(i, j)$, $p(m)$ is the number of pixels in the histogram of the image, and $L$ is the total number of grey levels in the image. The parameter $T$ is used to cluster the image into bright and dark regions, where the dark and bright regions have grey level ranges of $[0,T]$ and $[T+1,L-1]$, respectively.

### B. Histogram Segmentation

The original histogram is first divided into two parts with grey levels in the ranges of $[0,T]$ and $[T+1,L-1]$. Denote $N_{L1}$ and $N_{L2}$ as the total numbers of pixels in the grey level ranges $[0,T_l]$ and $[T_l+1,T]$, respectively. Denote $N_{U1}$ and $N_{U2}$ as the total numbers of pixels in the grey level ranges $[T+1,T_u]$ and $[T_u+1, L-1]$, respectively. $T_l$ and $T_u$ are grey levels that equally divide the grey level ranges $[0,T]$ and $[T+1, L-1]$ such that the number of pixels in the grey level ranges $[0,T_l]$ is equal to the number of pixels in $[T_l+1,T]$. Similarly for $[T+1,T_u]$ and $[T_u+1, L-1]$. Thus,

$$N_{L1} = N_{L2}, \quad (5)$$
$$N_{U1} = N_{U2}. \quad (6)$$

Finally, the input histogram is divided into four parts with grey level ranges of $[0,T_l]$, $[T_l+1,T]$, $[T+1,T_u]$ and $[T_u+1, L-1]$.

### C. Adaptive Histogram Clipping

Histogram bins clipping is introduced to prevent over-enhancement of local image areas. To limit the enhancement rate, it is necessary to limit the first derivative of the histogram or the histogram itself [11]. Clipping is performed if the histogram bins are greater than some threshold. For better enhancement, we introduce an adaptive method to clip the histogram. Specifically, the median of the occurring intensity of each of the corresponding sub-histogram are computed as

$$T_{L1}^m = median[h(k)], \quad \text{for } 0 \le k \le T_l, \quad (7)$$
$$T_{L2}^m = median[h(k)], \quad \text{for } T_l < k \le T, \quad (8)$$
$$T_{U1}^m = median[h(k)], \quad \text{for } T < k \le T_u, \quad (9)$$
$$T_{U2}^m = median[h(k)], \quad \text{for } T_u < k < L-1, \quad (10)$$

where $h(k)$ is the number of pixels with grey level $k$. The average number of grey level occurrences of each sub-histogram are computed as:

$$T_{L1}^a = \frac{1}{T_l+1} \sum_{k=0}^{T_l} h(k) \quad (11)$$

$$T_{L2}^a = \frac{1}{T-T_l} \sum_{k=T_l+1}^{T} h(k) \quad (12)$$

$$T_{U1}^a = \frac{1}{T_u-T} \sum_{k=T+1}^{T_u} h(k) \quad (13)$$

$$T_{U2}^a = \frac{1}{L-T_u-1} \sum_{k=T_u+1}^{L-1} h(k). \quad (14)$$

The clipping thresholds of the four sub-histograms are adaptively determined as follows:

$$T_{L1} = \begin{cases} T_{L1}^m & \text{if } T_{L1}^m > 0 \\ T_{L1}^a & \text{elsewhere} \end{cases} \quad (15)$$

$$T_{L2} = \begin{cases} T_{L2}^m & \text{if } T_{L2}^m > 0 \\ T_{L2}^a & \text{elsewhere} \end{cases} \quad (16)$$

$$T_{U1} = \begin{cases} T_{U1}^m & \text{if } T_{U1}^m > 0 \\ T_{U1}^a & \text{elsewhere} \end{cases} \quad (17)$$

$$T_{U2} = \begin{cases} T_{U2}^m & \text{if } T_{U2}^m > 0 \\ T_{U2}^a & \text{elsewhere} \end{cases}. \quad (18)$$

Finally, using the clipping thresholds, each sub-histogram is clipped as follows:

$$h_{L1}^N(k) = \begin{cases} h_{L1}(k) & \text{if } h_{L1}(k) \le T_{L1} \\ T_{L1} & \text{elsewhere} \end{cases} \quad (19)$$

$$h_{L2}^N(k) = \begin{cases} h_{L2}(k) & if\ h_{L2}(k) \leq T_{L2} \\ T_{L2} & elsewhere \end{cases} \quad (20)$$

$$h_{U1}^N(k) = \begin{cases} h_{U1}(k) & if\ h_{U1}(k) \leq T_{U1} \\ T_{U1} & elsewhere \end{cases} \quad (21)$$

$$h_{U2}^N(k) = \begin{cases} h_{U2}(k) & if\ h_{U2}(k) \leq T_{U2} \\ T_{U2} & elsewhere \end{cases}, \quad (22)$$

where $h_{L1}^N$, $h_{L2}^N$, $h_{U1}^N$, $h_{U2}^N$ are the number of pixels corresponding to the grey level of each clipped sub-histogram, respectively. The clipping of histograms is illustrated in Figure 1.

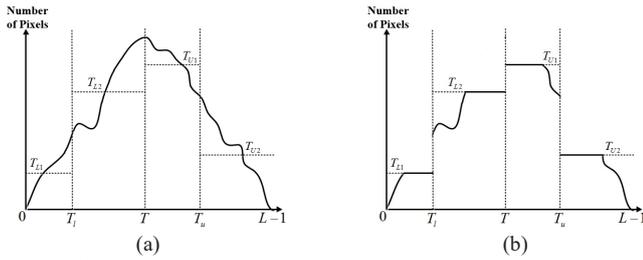

Fig. 1. Clipping of a histogram: (a) Original histogram and (b) the histogram after clipping.

### D. Equalization

After segmentation and clipping, four new sub-histograms are obtained. Histogram equalization is then applied to each sub-histogram as follows. Denote $N_{L1}^{new}$, $N_{L2}^{new}$, $N_{U1}^{new}$ and $N_{U2}^{new}$ as the total number of pixels in the new sub-histograms $L1$, $L2$, $U1$ and $U2$, respectively. The PDF of the corresponding sub-histograms are

$$P_{L1}(k) = \frac{h_{L1}^N(k)}{N_{L1}^{new}}, \quad for\ 0 \leq k \leq T_l \quad (23)$$

$$P_{L2}(k) = \frac{h_{L2}^N(k)}{N_{L2}^{new}}, \quad for\ T_{l+1} < k \leq T \quad (24)$$

$$P_{U1}(k) = \frac{h_{U1}^N(k)}{N_{U1}^{new}}, \quad for\ T < k \leq T_u \quad (25)$$

$$P_{U2}(k) = \frac{h_{U2}^N(k)}{N_{U2}^{new}}, \quad for\ T_u < k < L. \quad (26)$$

The cumulative density functions (CDF) of the corresponding sub-histograms are

$$C_{L1}(k) = \sum_{0}^{k} P_{L1}(k), \quad for\ 0 \leq k \leq T_l \quad (27)$$

$$C_{L2}(k) = \sum_{T_l+1}^{k} P_{L2}(k), \quad for\ T_l < k \leq T \quad (28)$$

$$C_{U1}(k) = \sum_{T+1}^{k} P_{U1}(k), \quad for\ T < k \leq T_u \quad (29)$$

$$C_{U2}(k) = \sum_{T_u+1}^{k} P_{U2}(k), \quad for\ T_u < k < L. \quad (30)$$

The next step of FIMHE is to equalize each sub-histogram using the transfer function

$$f(k) = \begin{cases} T_l \times [C_{L1}(k) - 0.5 \times P_{L1}(k)] & for\ 0 \leq k \leq T_l \\ (T - T_l - 1) \times [C_{L2}(k) - 0.5 \times P_{L2}(k)] + T_l + 1 & for\ T_l < k \leq T \\ (T_u - T - 1) \times [C_{U1}(k) - 0.5 \times P_{U1}(k)] + T + 1 & for\ T < k \leq T_u \\ (L - T_u - 2) \times [C_{U2}(k) - 0.5 \times P_{U2}(k)] + T_u + 1 & for\ T_u < k < L \end{cases}, \quad (31)$$

where $f(k)$ denotes the new grey levels of each sub-histogram after equalization. The enhanced image is generated from the combined four sub-histograms.

## IV. EXPERIMENTS

We performed qualitative and quantitative evaluations of the results of image enhancement, i.e., we evaluated FIMHE from the visual perspective of human perception and computed some quantitative assessments. We chose BBHE [6], DSIHE [7], RSIHE [10] (r=2), BHEPL [11], BHEPL-D [12], MHE [13] and ESIHE [14] for comparison because they are all based on a histogram segmentation, histogram modification or histogram clipping. We used the following four quantitative evaluation indices to evaluate the quality of the enhanced images: Shannon entropy (Entropy) [17]-[18], peak signal to noise ratio (PSNR) [19]-[22], absolute mean brightness error (AMBE) [23]-[26] and structural similarity index (SSIM) [27]-[33].

Entropy is an index that can be used to determine the richness of image details and has been widely used in the evaluation of image quality. Entropy can be defined as:

$$Entropy = -\sum_{k=0}^{L-1} p(k) \log_2 p(k), \quad (32)$$

where $p(k)$ is the PDF of an image histogram. A larger entropy indicates richer information. For comparison, we also compute the percentage of information entropy, i.e.,

$$Entropy\% = \frac{Entropy_{outputimage}}{Entropy_{inputimage}} \times 100\%. \quad (33)$$

PSNR measures image distortion or noise level and is defined as:

$$PSNR = 10 \log_{10} \left[ \frac{(L-1)^2}{MSE} \right], \quad (34)$$

where the mean squared error (MSE) [34]-[36] is given by

$$MSE = \frac{1}{M \times N} \sum_{i=1}^{M} \sum_{j=1}^{N} |X(i,j) - Y(i,j)|^2. \quad (35)$$

$X(i,j)$ and $Y(i,j)$ are respectively the grey level of the original image and enhanced image at coordinates $(i,j)$.

AMBE utilizes the absolute value of the difference in the mean brightness of the original image and the enhanced image, i.e.,

$$AMBE = |E(X) - E(Y)|, \quad (36)$$

where

$$E(X) = \frac{1}{M \times N} \sum_i \sum_j X(i,j) \quad (37)$$

$$E(Y) = \frac{1}{M \times N} \sum_i \sum_j Y(i,j). \quad (38)$$

It is used to evaluate the change in brightness before and after image enhancement. Ideally, the mean brightness of the enhanced image should be equal to the mean brightness of the input image, in which case the value of AMBE is zero [37].

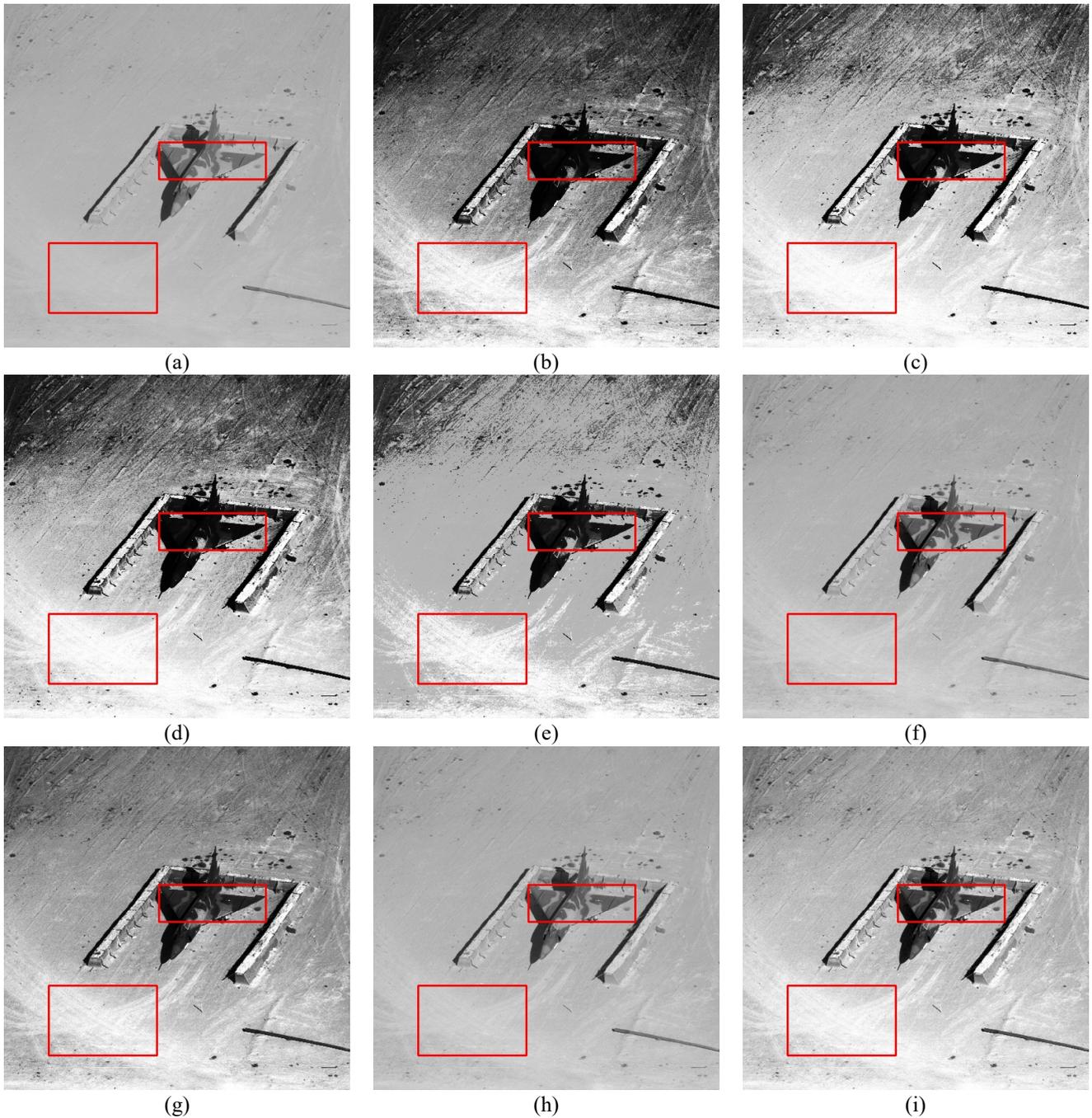

Fig. 2. Plane. (a) Original image. From (b) to (i): enhanced images using BBHE, DSIHE, RSIHE, BHEPL, BHEPL-D, MHE, ESIHE, and the proposed method, respectively.

SSIM combines correlation distortion, image brightness distortion and contrast distortion to determine the degree of distortion, i.e.,

$$SSIM(x,y) = \frac{(2\mu_x\mu_y + C_1)(2\sigma_{xy} + C_2)}{(\mu_x^2 + \mu_y^2 + C_1)(\sigma_x^2 + \sigma_y^2 + C_2)}, \quad (39)$$

where $\mu_x$ and $\mu_y$ are respectively the average brightness of the original image and the enhanced image; $\sigma_x$ and $\sigma_y$ are respectively the standard deviation of the original image and the enhanced image; $\sigma_{xy}$ is the square root of the covariance of the original image and the enhanced image; and $C_1$ and $C_2$ are constants. The range of SSIM is [0, 1]. In order to get a better-enhanced image, the value of SSIM should be larger. The larger the value the smaller is the image distortion.

### A. Qualitative Evaluation

In order to analyze the quality of the enhanced image from visual effects, we selected six groups of images "Plane", "Photographer", "House", "Fish", "Couple" and "U2" which are commonly used in image quality assessment. The obvious changes in the enhanced images are denoted with red boxes in Figures 2 - 7.

For the first image "Plane" in Figure 2(a), we note that the texture features of the original image are blurred and many details are obscured. As can be seen in the rest of Figure 2, the results of BBHE and RSIHE are obviously over enhanced and have a poor performance. Although MHE has enhanced the texture slightly, the enhanced image is too dark on the wing and the foreground. The image content of the result of our proposed method is the most abundant and has a smooth texture.

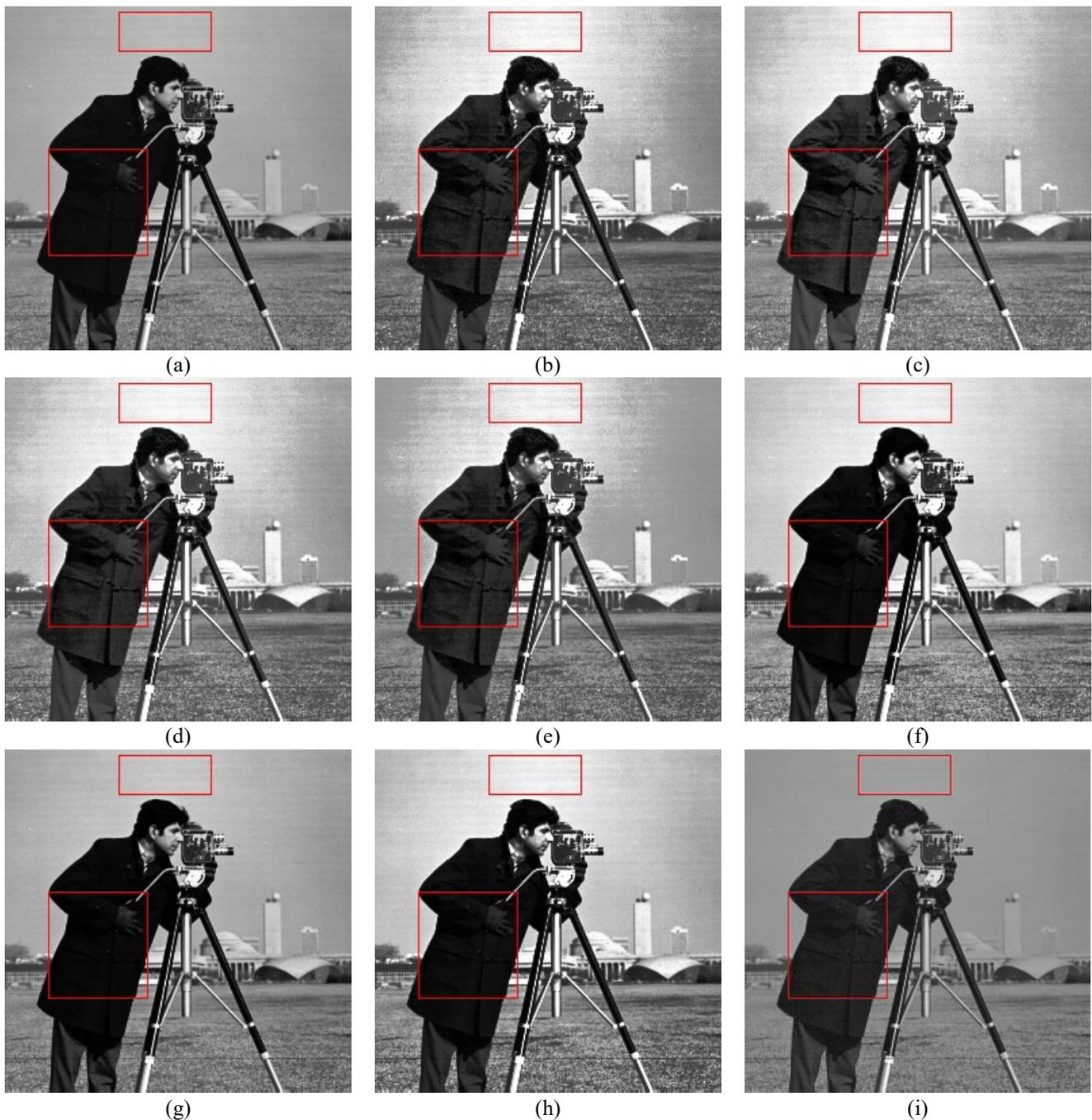

Fig. 3. Photographer. (a) Original image. From (b) to (i): enhanced images using BBHE, DSIHE, RSIHE, BHEPL, BHEPL-D, MHE, ESIHE, and the proposed method, respectively.

Figure 3(a) shows the image "Photographer" which is too dark for the image details to be clearly visible. The rest of Figure 3 shows that using DSIHE and BHEPL resulted in a saturation effect that makes the enhanced image greatly different in brightness from the original image. The result of the proposed method has a more natural look compared with those of the other methods, and the brightness is the most similar to the original image. It can also be observed from the man's clothes that the details of the original image are well preserved and has the minimum image degradation.

For the image "House" shown in Figure 4(a), DSIHE and RSIHE introduced a high level of noise and caused serious image distortion. Unlike the proposed method, the other methods produced excessive enhancement to the window, and resulted in a loss of image details. The proposed method not only resulted in the most abundant image details, the average brightness of the enhanced image is the most similar to the original image.

For the image "Fish" shown Figure 5(a), the proposed method resulted in the most natural image with its brightness basically the same as the original image. The enhanced image has the clearest outline and most vivid details especially when compared with the result of BHEPL, which looks unnaturally artistic. It also has the richest details in the image while minimizing the noise level.

Figure 6(a) shows the image "Couple" which is too dark for the image details to be clearly visible. Using DSIHE and MHE resulted in a saturation effect that makes the enhanced image greatly different from the original image in brightness. The result of the proposed method has a more natural look compared with those of the other methods, and the brightness is the most similar to that in the original image. It can also be observed from the mural and the man's pants in the enhanced image that the details of the original image are well preserved and has the minimum image degradation.

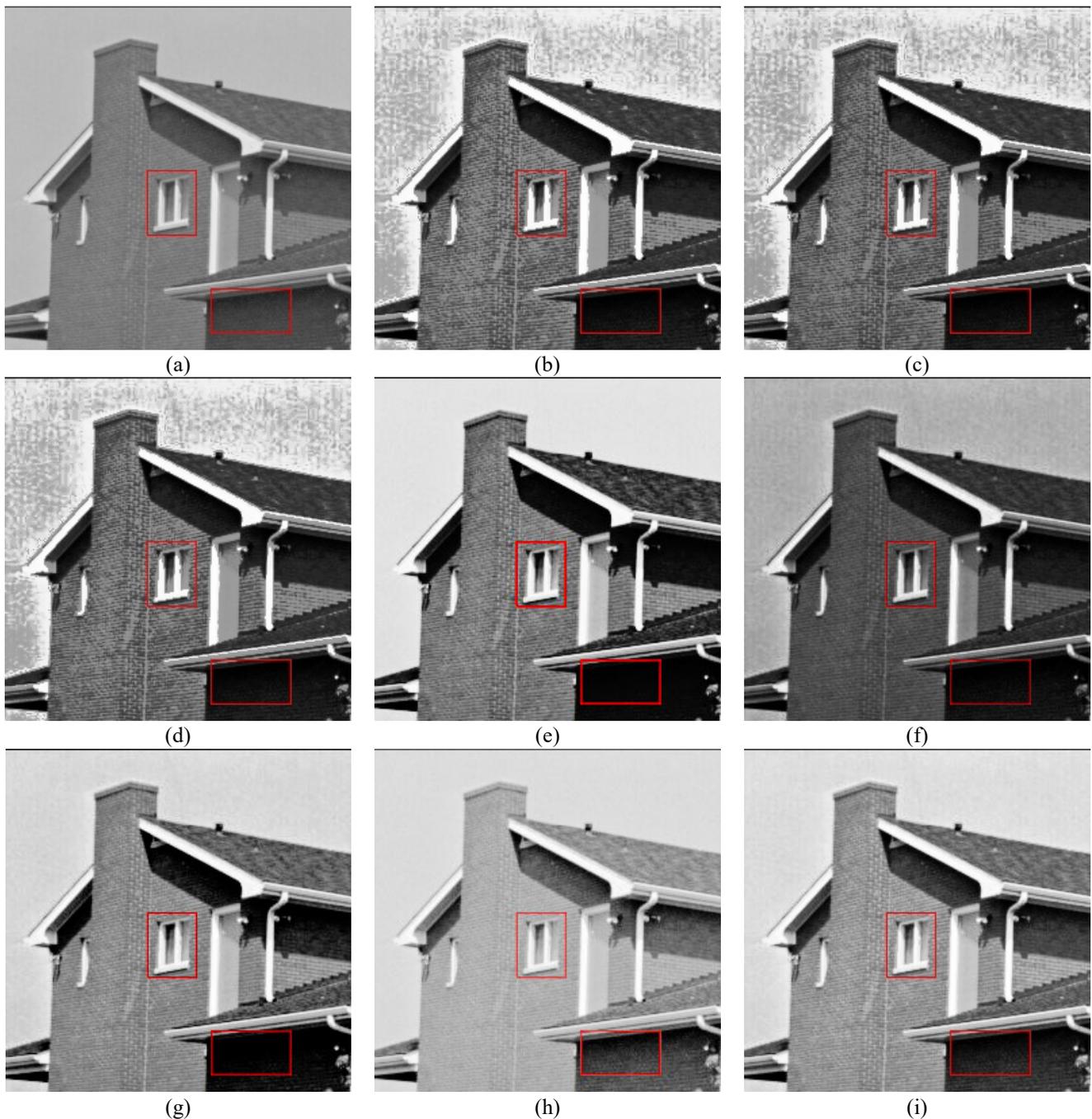

Fig. 4. House. (a) Original image. From (b) to (i): enhanced images using BBHE, DSIHE, RSIHE, BHEPL, BHEPL-D, MHE, ESIHE, and the proposed method, respectively.

For the image "U2" shown in Figure 7(a), DSIHE and ESIHE introduced a high level of noise and caused serious image distortion. Unlike the proposed method, other methods produced excessive enhancement to the left wing, and resulted in a loss of image details. The proposed method not only resulted in the most abundant image details, the average brightness of the image is the most similar to the original image.

Histogram equalization refers to processing an input image by mapping the input into an output image to effectively utilize the dynamic range such that each grey level has an equal number of pixels in the output. Therefore, it is desirable for the equalized output image to have a flat grey level distribution. However, it should be noted that this process should not change the overall shape of the input histogram so as to preserve the image content.

Figure 8 shows the histograms of the test images after applying different image enhancement algorithms. Due to the space limitations, only a part of them are shown. The original histogram is stretched to both ends of the grey scale. As can be seen from the figure, each algorithm has improved the histogram of the original image. However, most algorithms greatly change the shape of the original histogram, this means that the information of the image has been changed, resulting in a large difference in visual appearance between the original and enhanced images. The histogram of the image processed by FIMHE has greatly flattened grey-level distribution. This algorithm stretches the original histogram to both ends of the grey scale while maintaining the shape of the original histogram. It thus preserves the image details and produces visual experience most similar to that of the original image.

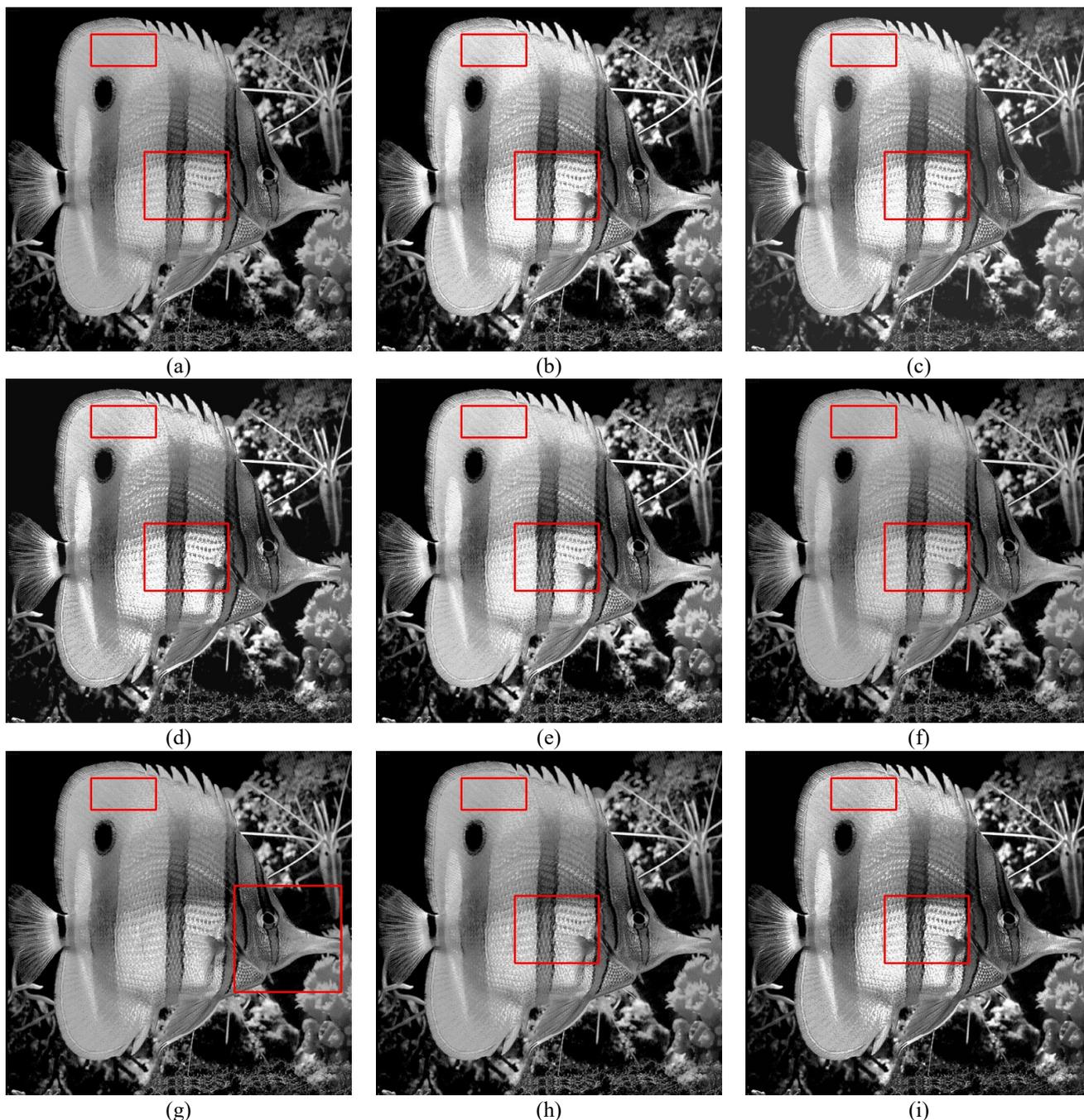

Fig. 5. Fish. (a) Original image. From (b) to (i): enhanced images using BBHE, DSIHE, RSIHE, BHEPL, BHEPL-D, MHE, ESIHE, and the proposed method, respectively.

*B. Quantitative Evaluation*

Tables I to VI show the quantitative evaluation results on "Plane", "Photographer", "House", "Fish", "Couple" and "U2", respectively. The best test results are denoted in bold.

Table I shows that the proposed method, FIMHE, performs better on "Plane" than the other seven algorithms on all evaluation indices. In particular, its AMBE value is the smallest, i.e., the brightness of the original image is best preserved. Although the AMBE value of the image enhanced using BBHE is similar to that of FIMHE, it performs poorly on other indices.

Table II shows that FIMHE outperforms the other algorithms on "Photographer". Although its Entropy is slightly smaller than the BHEPL-D, its PSNR and SSIM are maximum, i.e., the image information is abundant with minimal noise and artifacts. Although its Entropy is slightly smaller than that of the original image, FIMHE is still better than the other seven algorithms.

Table III shows that FIMHE is much better for Entropy and PSNR on "House". The brightness of the original image is preserved, and the introduction of noise is minimal, ensuring that the enhanced image can be used for televisions and digital cameras. The good effects of FIMHE can also be observed by comparing the details of the images shown in Figure 4.

Table IV shows the proposed method produces the largest

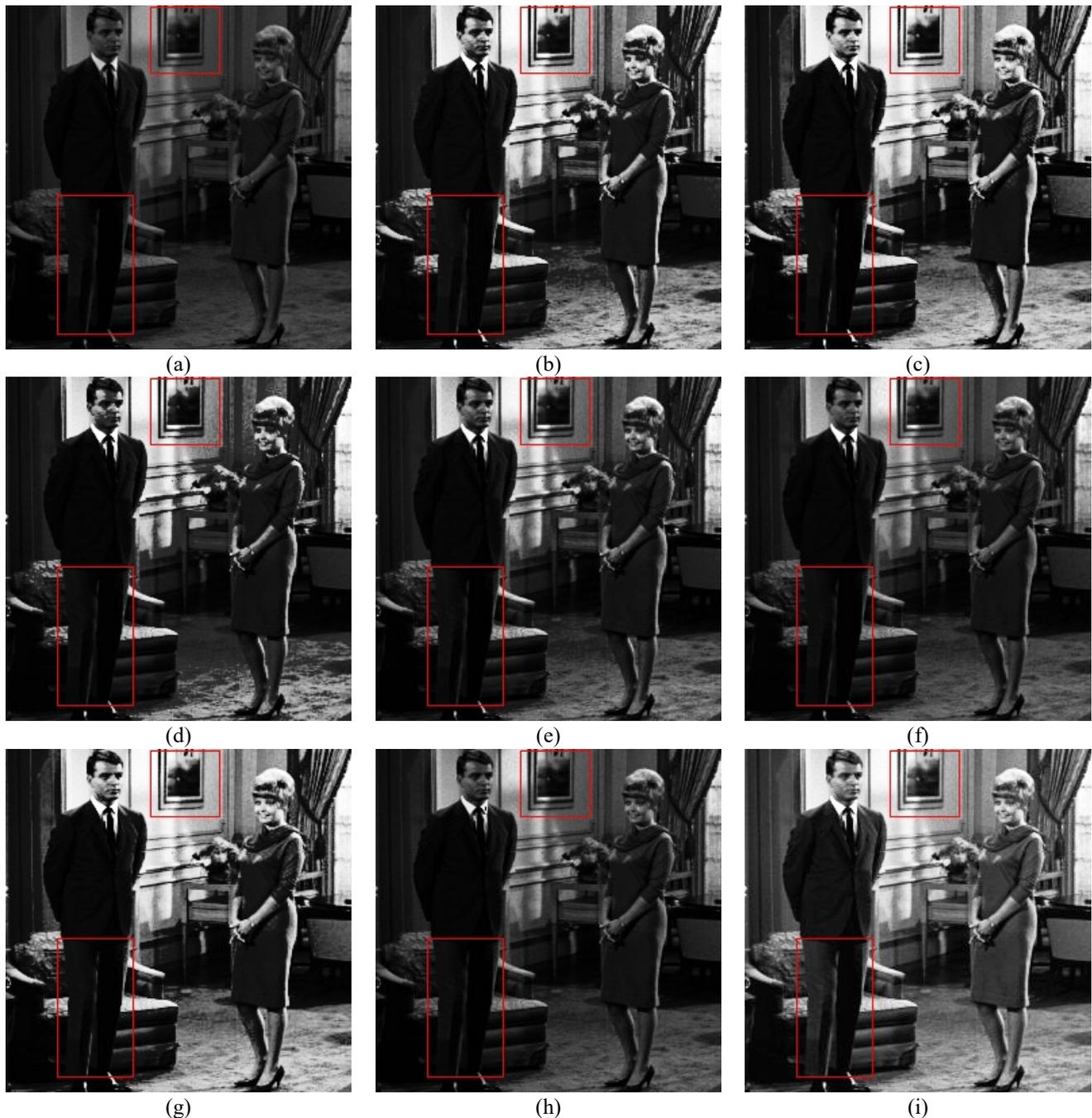

Fig. 6. Couple. (a) Original image. From (b) to (i): enhanced images using BBHE, DSIHE, RSIHE, BHEPL, BHEPL-D, MHE, ESIHE, and the proposed method, respectively.

SSIM value on "Fish" compared with other methods. This means that the distortion of the image is minimal, which gives the enhanced image a more natural look.

Table V shows that FIMHE is much better for PSNR and AMBE on "Couple". The brightness of the original image is preserved, and the introduction of noise is minimal, ensuring that the enhanced image can be used for televisions and digital cameras.

Table VI shows the proposed method produces the largest SSIM and PSNR on "U2" (i.e., achieving the best performance) compared with the other methods. As can be seen from the maximum AMBE, it is closest to the brightness of the original image. It can also be found that the image processed by the proposed method achieves the highest (i.e., the best) value for SSIM and PSNR. This means that the distortion of the image is minimal, giving the enhanced image a more natural look.

The performance of FIMHE is more satisfactory than the other seven algorithms on "Plane", "Photographer", "House", "Fish", "Couple" and "U2".

In addition to these six test images, four objective evaluation functions (i.e., Entropy, PSNR, AMBE, and SSIM) were employed on different image databases to further verify the capabilities and performance of the proposed FIMHE.

Figure 9 shows the average test results for 400 images in the Berkeley database. It shows that FIMHE obtains the maximum average Entropy, i.e., the information of the original image is retained to the greatest degree, and the

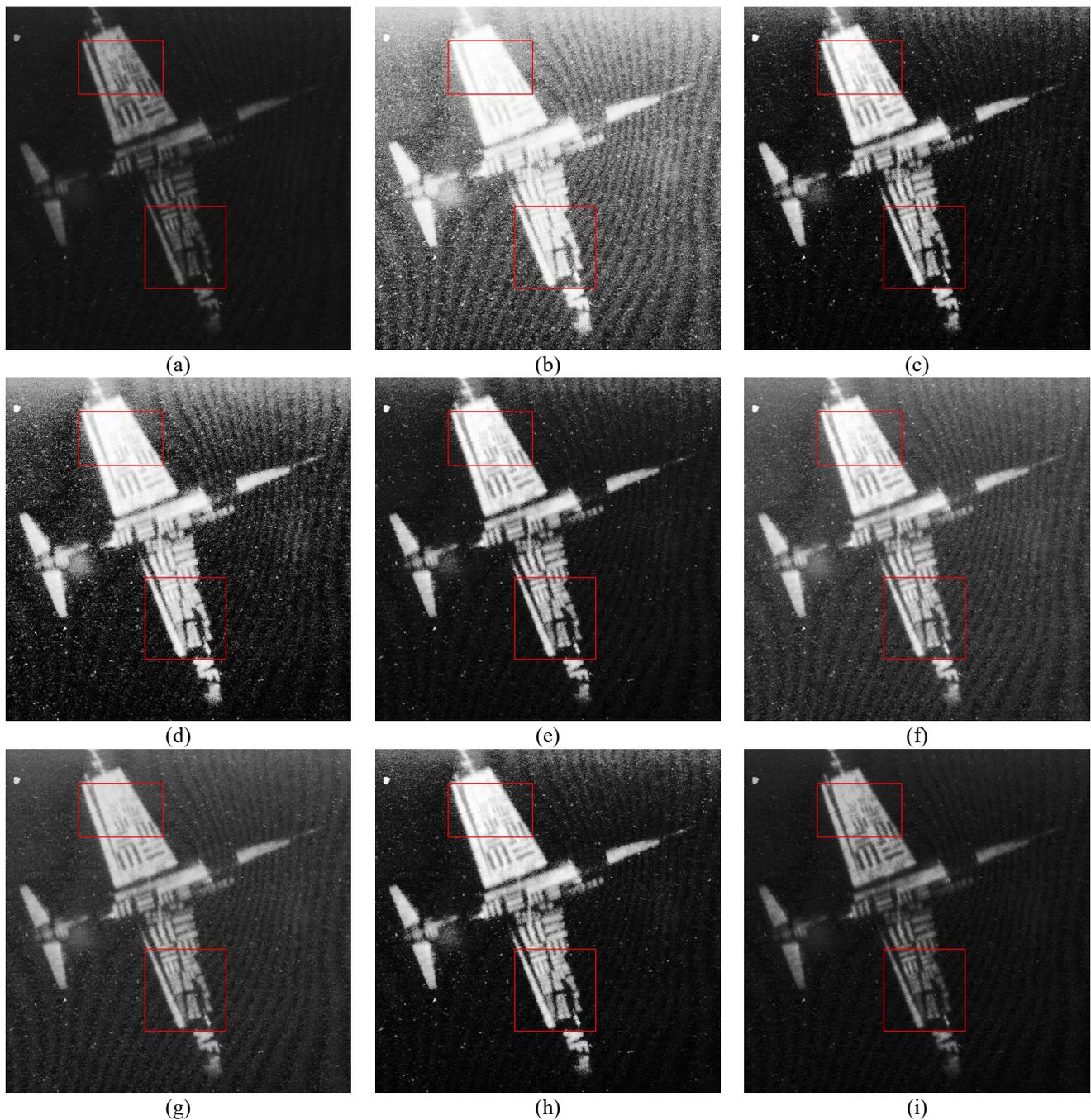

Fig. 7. U2. (a) Original image. From (b) to (i): enhanced images using BBHE, DSIHE, RSIHE, BHEPL, BHEPL-D, MHE, ESIHE, and the proposed method, respectively.

TABLE I
THE RESULT OF DIFFERENT ALGORITHMS ON THE IMAGE "PLANE"

| Methods | Entropy | Entropy% | PSNR | AMBE | SSIM |
|---|---|---|---|---|---|
| Original image | 4.938 | - | - | - | - |
| BBHE | 4.813 | 97.168 | 13.310 | 0.937 | 0.459 |
| DSIHE | 4.785 | 96.901 | 12.127 | 16.736 | 0.387 |
| RSIHE | 4.732 | 95.835 | 15.897 | 7.183 | 0.518 |
| BHEPL | 4.914 | 99.528 | 19.159 | 8.641 | 0.705 |
| BHEPL-D | 4.910 | 99.432 | 29.511 | 1.449 | 0.959 |
| MHE | 4.891 | 99.060 | 15.991 | 16.921 | 0.549 |
| ESIHE | 4.893 | 99.104 | 20.332 | 9.706 | 0.740 |
| FIMHE | **4.931** | **99.873** | **31.606** | **0.916** | **0.965** |

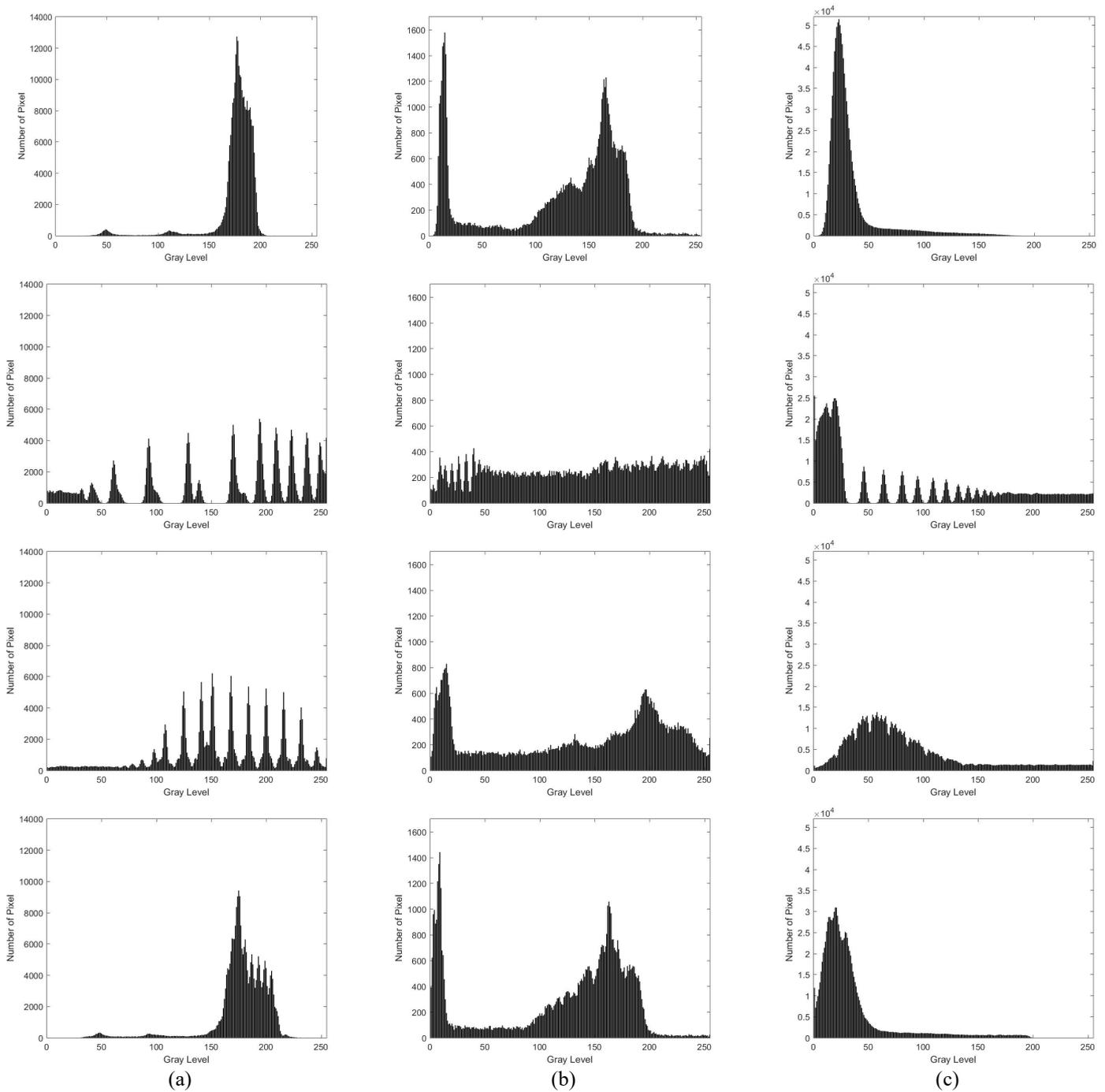

Fig. 8. Top row: Histogram of test image (a) Plane; (b) Photographer; and (c) U2. From row 2 to bottom each row respectively shows the result of the images processed by DSIHE, MHE, and FIMHE.

TABLE II
THE RESULT OF DIFFERENT ALGORITHMS ON THE IMAGE "PHOTOGRAPHER"

| Methods | Entropy | Entropy% | PSNR | AMBE | SSIM |
|---|---|---|---|---|---|
| Original image | 7.019 | - | - | - | - |
| BBHE | 6.809 | 97.008 | 18.269 | 23.678 | 0.820 |
| DSIHE | 6.781 | 96.609 | 18.778 | 17.388 | 0.817 |
| RSIHE | 6.765 | 96.381 | 21.166 | 8.323 | 0.824 |
| BHEPL | 6.922 | 98.618 | 18.864 | 19.332 | 0.919 |
| BHEPL-D | **6.996** | **99.672** | 30.708 | 1.634 | 0.964 |
| MHE | 6.901 | 98.319 | 19.566 | 14.525 | 0.911 |
| ESIHE | 6.900 | 98.305 | 20.000 | 11.754 | 0.908 |
| FIMHE | 6.967 | 99.259 | **36.275** | **1.609** | **0.965** |

TABLE III
THE RESULT OF DIFFERENT ALGORITHMS ON THE IMAGE "HOUSE"

| Methods | Entropy | Entropy% | PSNR | AMBE | SSIM |
|---|---|---|---|---|---|
| Original image | 6.488 | - | - | - | - |
| BBHE | 6.246 | 96.270 | 17.267 | 13.920 | 0.562 |
| DSIHE | 6.242 | 96.208 | 17.589 | 9.910 | 0.567 |
| RSIHE | 6.240 | 96.178 | 20.815 | **0.816** | 0.717 |
| BHEPL | 6.424 | 99.014 | 21.824 | 7.186 | 0.886 |
| BHEPL-D | 6.426 | 99.044 | 23.818 | 1.565 | 0.935 |
| MHE | 6.393 | 98.536 | 18.506 | 18.565 | 0.856 |
| ESIHE | 6.410 | 98.798 | 20.479 | 7.378 | 0.854 |
| FIMHE | **6.446** | **99.353** | **29.492** | 2.582 | **0.984** |

TABLE IV
THE RESULT OF DIFFERENT ALGORITHMS ON THE IMAGE "FISH"

| Methods | Entropy | Entropy% | PSNR | AMBE | SSIM |
|---|---|---|---|---|---|
| Original image | 6.451 | - | - | - | - |
| BBHE | 6.131 | 95.031 | 18.969 | 24.759 | 0.724 |
| DSIHE | 6.138 | 95.137 | 19.060 | 24.124 | 0.721 |
| RSIHE | 6.109 | 94.698 | 23.904 | 7.746 | 0.779 |
| BHEPL | 6.399 | 99.188 | 23.052 | 11.587 | 0.952 |
| BHEPL-D | 6.407 | 99.311 | 24.285 | 9.832 | 0.959 |
| MHE | 6.318 | 97.933 | 22.425 | 12.005 | 0.909 |
| ESIHE | 6.299 | 97.646 | 26.152 | 0.857 | 0.931 |
| FIMHE | **6.412** | **99.386** | **43.513** | **0.514** | **0.998** |

TABLE V
THE RESULT OF DIFFERENT ALGORITHMS ON THE IMAGE "COUPLE"

| Methods | Entropy | Entropy% | PSNR | AMBE | SSIM |
|---|---|---|---|---|---|
| Original image | 6.473 | - | - | - | - |
| BBHE | 6.306 | 97.413 | 13.265 | 32.275 | 0.693 |
| DSIHE | 6.267 | 96.807 | 11.667 | 42.987 | 0.620 |
| RSIHE | 6.339 | 97.924 | 15.699 | 19.435 | 0.802 |
| BHEPL | 6.373 | 98.444 | 18.598 | 15.058 | 0.879 |
| BHEPL-D | 6.412 | 99.052 | 22.268 | 10.915 | 0.930 |
| MHE | 6.410 | 99.022 | 11.681 | 53.587 | 0.489 |
| ESIHE | 6.410 | 99.019 | 14.422 | 38.979 | 0.614 |
| FIMHE | **6.440** | **99.490** | **23.759** | **8.686** | **0.950** |

TABLE VI
THE RESULT OF DIFFERENT ALGORITHMS ON THE IMAGE "U2"

| Methods | Entropy | Entropy% | PSNR | AMBE | SSIM |
|---|---|---|---|---|---|
| Original image | 5.650 | - | - | - | - |
| BBHE | 5.558 | 98.375 | 15.077 | 16.292 | 0.592 |
| DSIHE | 5.485 | 97.069 | 10.975 | 41.301 | 0.303 |
| RSIHE | 5.535 | 97.961 | 15.646 | 17.117 | 0.668 |
| BHEPL | 5.570 | 98.583 | 21.007 | 3.297 | 0.854 |
| BHEPL-D | 5.541 | 98.072 | 23.411 | 5.068 | 0.903 |
| MHE | 5.601 | 99.125 | 12.414 | 51.345 | 0.387 |
| ESIHE | 5.601 | 99.133 | 15.812 | 33.704 | 0.551 |
| FIMHE | **5.626** | **99.567** | **28.800** | **0.683** | **0.920** |

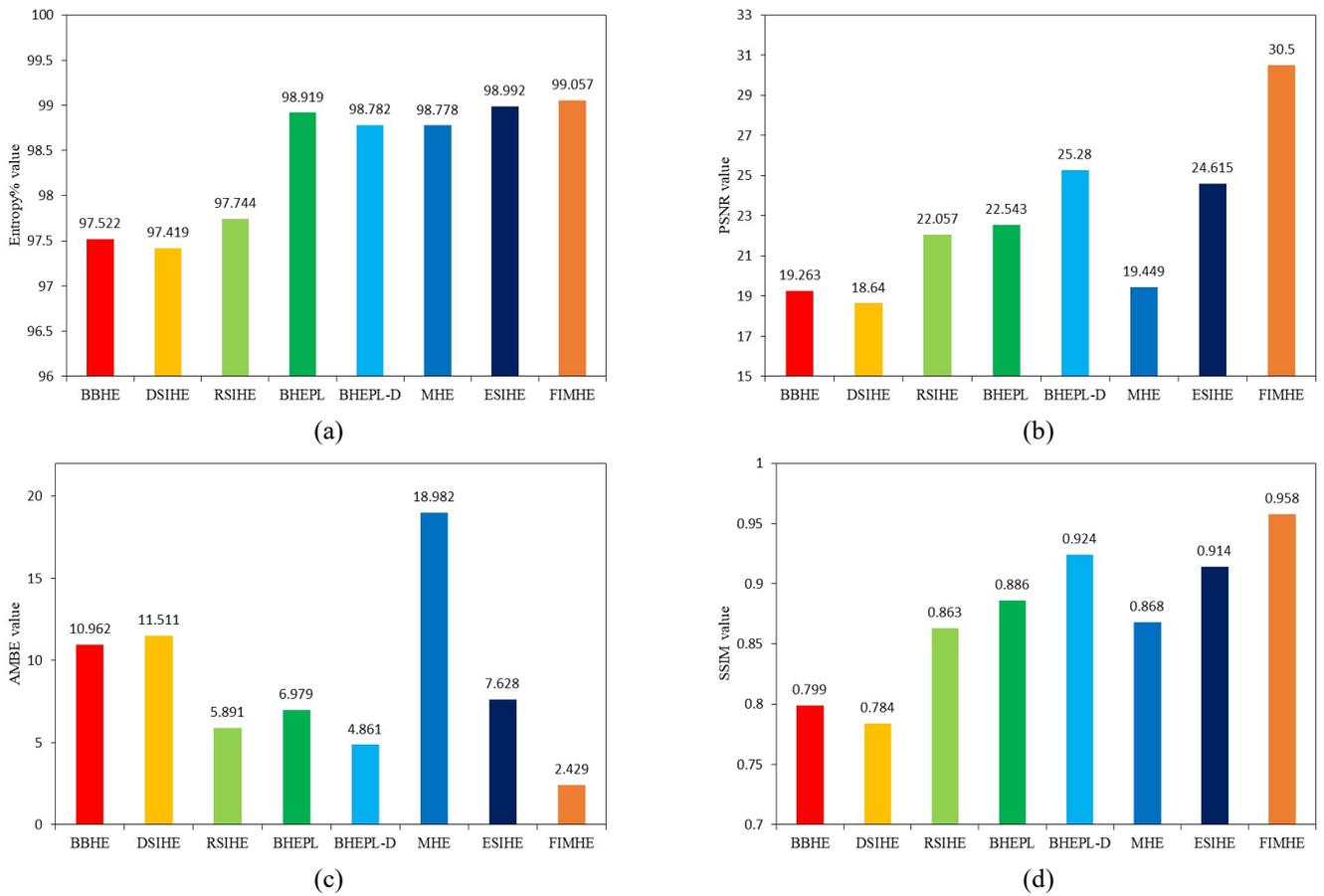

Fig. 9. Average result of (a) Entropy, (b) PSNR, (c) AMBE and (d) SSIM for the 400 images in the BERKELEY Database.

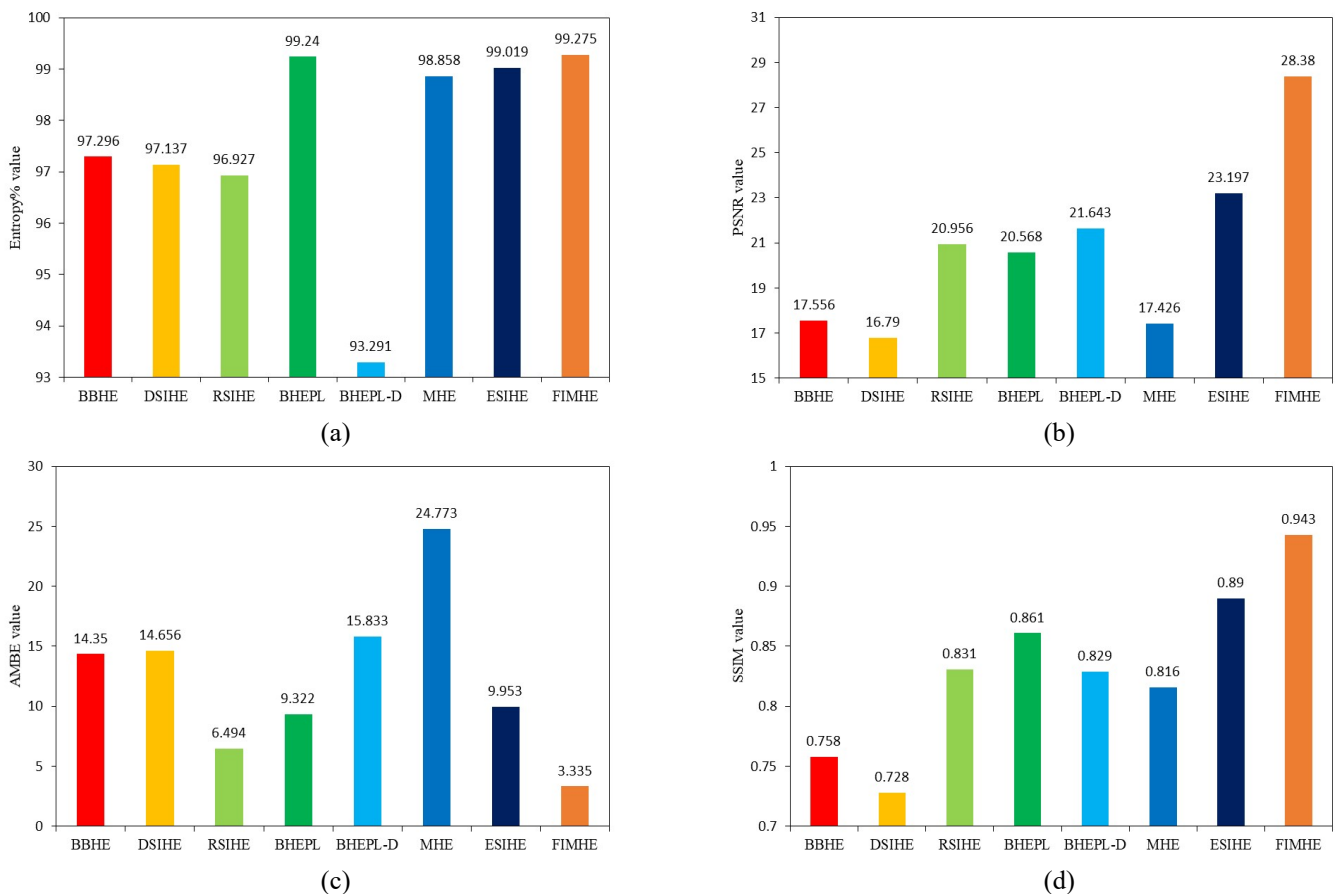

Fig. 10. Average result of (a) Entropy, (b) PSNR, (c) AMBE and (d) SSIM for the 86 images in the CVF-UGR-IMAGE Database.

details are fully displayed. Its PSNR is the highest, i.e., the image contrast is well enhanced while introducing the smallest noise level. Its average AMBE is the smallest, i.e., the average brightness of the original image is best preserved. Finally, its SSIM being the highest indicates that the proposed method produces the minimum image distortion.

Figure 10 shows the same conclusion on the CVF-UGR-Image database. Therefore, the proposed method is better than the state-of-the-art HE-based image enhancement methods.

## V. Conclusion

This paper proposes a HE-based enhancement algorithm, FIMHE, which combines fuzzy intensity measure, image segmentation and adaptive clipping of histogram bins to address over-enhancement and image distortion in HE-based enhanced images. The algorithm not only enhances the contrast of the original image, but also preserves the original information in the image, making the enhanced image more natural in appearance. FIMHE avoids the problems of excessive enhancement, high level of noise and severe image distortion, which are present in images enhanced by other HE-based methods. When compared with the state-of-the-art HE-based contrast enhancement methods, FIMHE preserves the texture details and retains the overall content of the original image while providing sufficient contrast. The enhanced images also have the most natural appearance. The proposed algorithm is simple but effective for image contrast enhancement. It operates in real time, and has low requirements on equipment which reduces its operational cost. Therefore, it can be applied in real-time applications that require image contrast enhancement while preserving the overall image content.

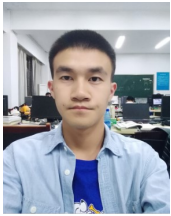

**Xiangyuan Zhu** received his BS degree in automation from Xiangtan University, China, in 2017. He is currently pursuing his MS degree in control science and engineering at Central South University, China. His research interests include computer vision and artificial intelligence.

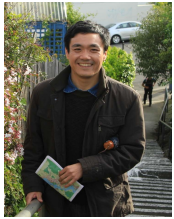

**Xiaoming Xiao** received his BS and MS degrees in engineering from Central South University China in 1985 and 1988, respectively, and his PhD degree in control science and engineering in 2000 from Central South University China. He has been an associate professor at Central South University China since 2002. His research interests include computer vision and intelligent control.

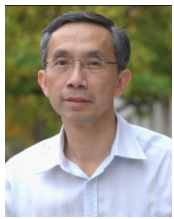

**Tardi Tjahjadi** received his BS degree in from University College London in 1980, his MS degree in management from UMIST, U.K., in 1981, and his PhD degree in total technology from UMIST. He has been an Associate Professor with The University of Warwick since 2000 and a Reader since 2014. His research interests include image processing and computer vision.

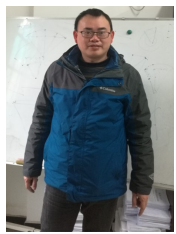

**Zhihu Wu** received his BS and MS degrees in control science and engineering from Harbin Institute of Technology in 2002 and 2004, respectively, and his PhD degree in control science and engineering in 2009 from Harbin Institute of Technology China. He has been a lecturer at Central South University China Since 2009. His research interests include embedded system, computer vision and automotive control.

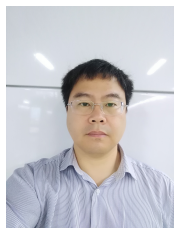

**Jin Tang** received his BS degree in mechanics and his MS degree from Peking University, China, in 1987 and 1990, respectively, and his PhD degree in pattern recognition and intelligence system from Central South University, China, in 2002. He has been a Professor with Central South University since 2005. His research interests include computer vision and artificial intelligence.